\documentclass[
]
{llncs}
\usepackage{makeidx}

\usepackage[textsize=scriptsize
]{todonotes}
\usepackage{tabu}
\usepackage{multirow}
\usepackage{graphicx}
\usepackage{booktabs}
\usepackage{tabu}
\usepackage{subcaption}
\usepackage[font=small]{caption}
\usepackage{blindtext}
\usepackage{amsmath}
\usepackage{hyperref}
\usepackage{url}

\usepackage{siunitx,etoolbox}
\robustify\bfseries
\usepackage[capitalize]{cleveref}

\makeatletter
\DeclareRobustCommand\onedot{\futurelet\@let@token\@onedot}
\def\@onedot{\ifx\@let@token.\else.\null\fi\xspace}

\def\eg{\emph{e.g}\onedot} 
\def\ie{\emph{i.e}\onedot} 
\def\cf{\emph{cf.}\xspace}

\makeatother

\newcommand{\Dtoy}{\mbox{\textsc{Toy}}\xspace}
\newcommand{\Dtragen}{\mbox{\textsc{TRAgen}}\xspace}
\newcommand{\Ddsb}{\mbox{\textsc{DSB2018}}\xspace}

\newcommand{\stardist}{\mbox{\small\textsc{StarDist}}\xspace}

\begin{document}

\title{Cell Detection with Star-convex Polygons}

\author{Uwe Schmidt\inst{1,\star}, Martin Weigert\inst{1,\star}, Coleman Broaddus\inst{1}, \and Gene Myers\inst{1,2}}
\institute{
  Max Planck Institute of Molecular Cell Biology and Genetics, Dresden, Germany
 \newline
 Center for Systems Biology Dresden, Germany
  \and
  Faculty of Computer Science, Technical University Dresden, Germany
}

\maketitle

\let\oldthefootnote\thefootnote
\renewcommand{\thefootnote}{\fnsymbol{footnote}}
\footnotetext[1]{Equal contribution.}
\let\thefootnote\oldthefootnote

\begin{abstract}
  Automatic detection and segmentation of cells and nuclei in microscopy images is important for many biological applications. Recent successful learning-based approaches include per-pixel cell segmentation with subsequent pixel grouping, or localization of bounding boxes with subsequent shape refinement. In situations of crowded cells, these can be prone to segmentation errors, such as falsely merging bordering cells or suppressing valid cell instances due to the poor approximation with bounding boxes.
To overcome these issues, we propose to localize cell nuclei via \emph{star-convex polygons}, which are a much better shape representation as compared to bounding boxes and thus do not need shape refinement. To that end, we train a convolutional neural network that predicts for every pixel a polygon for the cell instance at that position.
We demonstrate the merits of our approach on two synthetic datasets and one challenging dataset of diverse fluorescence microscopy images.

\end{abstract}

\newcommand{\figOverview}{{
\begin{figure}[t]
  \centering
  {\includegraphics[width=1\textwidth]{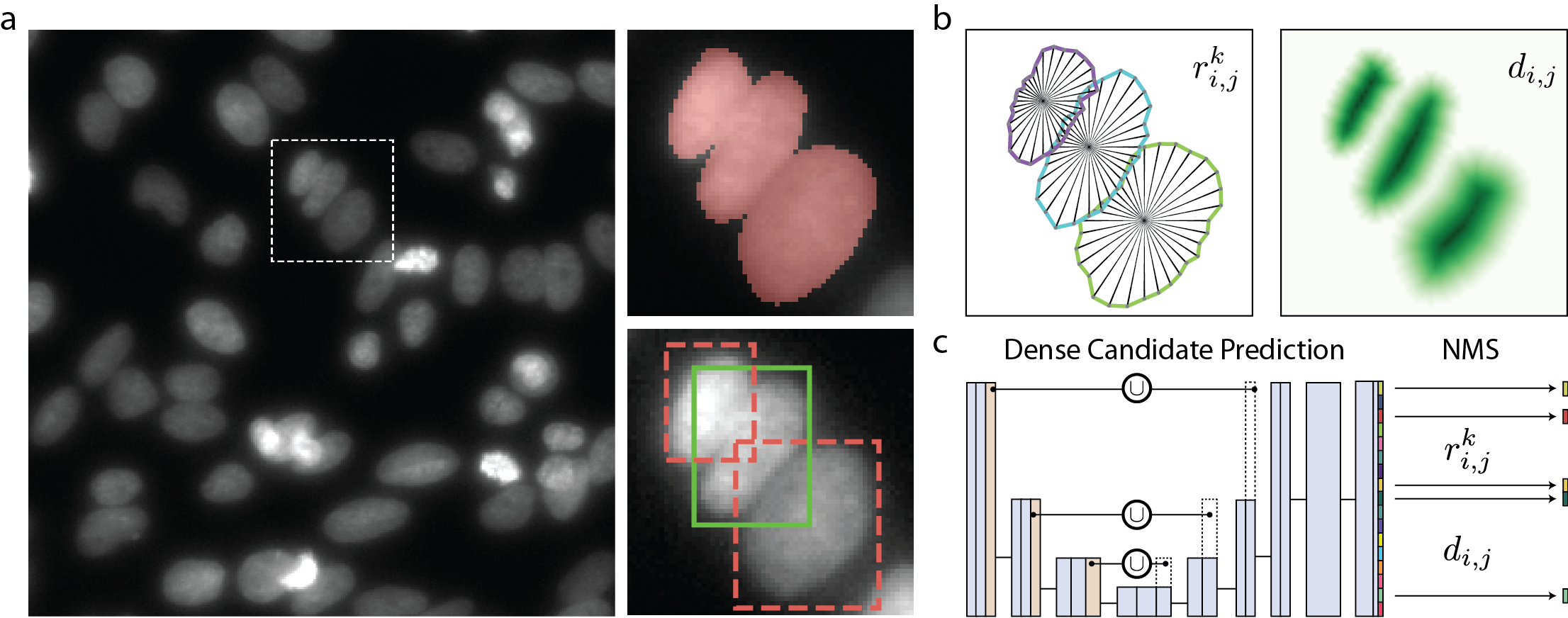}}%
  \vspace{-3pt}
\caption{%
\emph{(a)} Potential segmentation errors for images with crowded nuclei: Merging of touching cells (upper right) or suppression of valid cell instances due to large overlap of bounding box localization (lower right).
\emph{(b)} The proposed \stardist method predicts object probabilities $d_{i,j}$ and star-convex polygons parameterized by the radial distances $r^k_{i,j}$.
\emph{(c)} We densely predict $r^k_{i,j}$ and $d_{i,j}$ using a simple U-Net architecture~\cite{ronneberger2015} and then select the final instances via non-maximum suppression (NMS).
}
\label{fig:overview}
\end{figure}
}}

\newcommand{\figPolyExample}{{
\begin{figure}[b]
\vspace{-6pt}
  \centering
  {\includegraphics[width=1\textwidth,clip,trim=0mm 5mm 0mm 0mm]{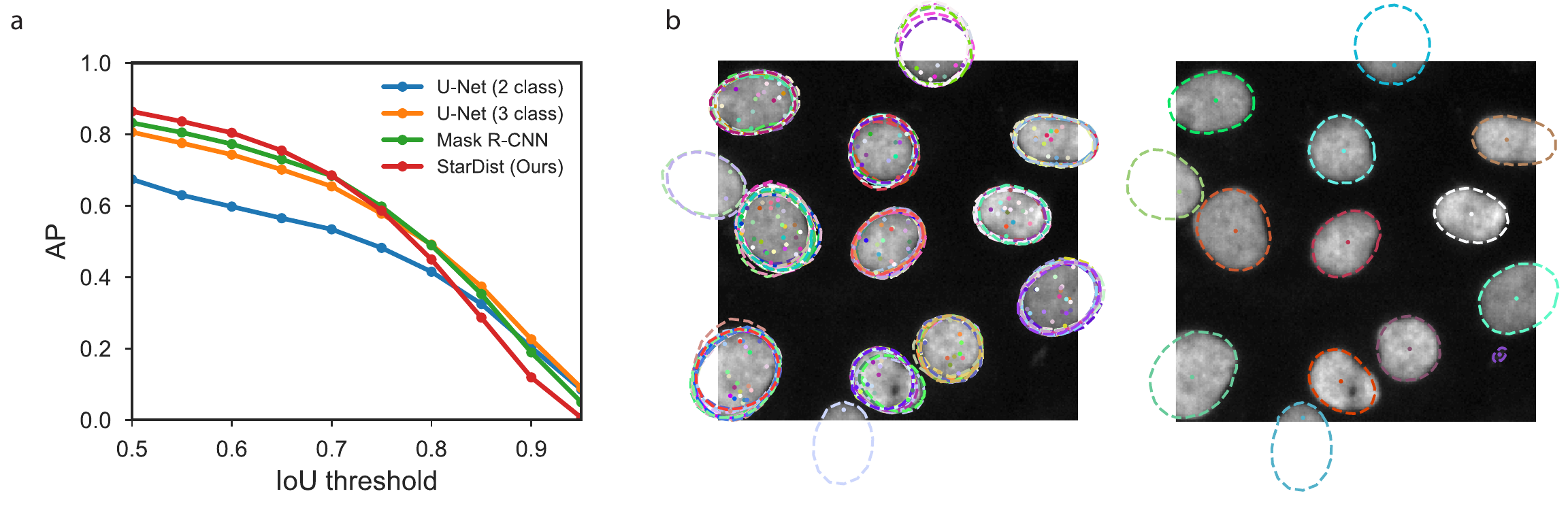}}%
\vspace{-6pt}
\caption{%
\emph{(a)} Detection scores on dataset \Ddsb (\cf\cref{tab:results}, bottom).
\emph{(b)} Example of \stardist polygon predictions for $200$ random pixels (left)
and for all pixels after non-maximum suppression (right); pixels and associated polygons are color-matched.
}
\label{fig:poly_example}
\vspace{-6pt}
\end{figure}
}}

\newcommand{\figResultsA}{{
\begin{figure}[t!]
  \centering
  {\includegraphics[width=1\textwidth,clip,trim=10mm 8mm 2mm 2mm]{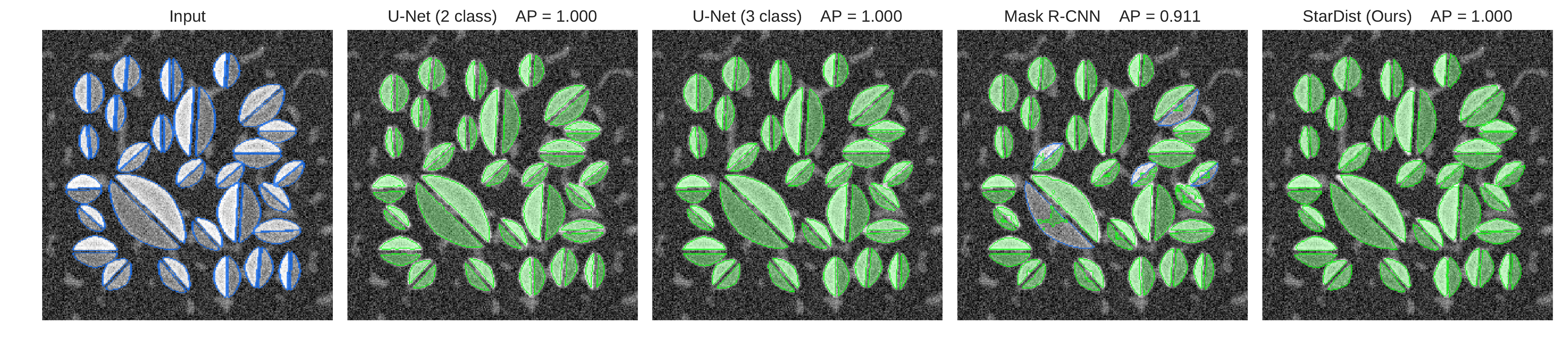}}%
\vspace{-8pt}
\caption{Segmentation result ($\tau=0.5$) for \Dtoy image.
Predicted cell instances are depicted in green if correctly matched (\emph{TP}), otherwise highlighted in red (\emph{FP}).
Ground truth cells are always shown by their blue outlines in the input image (left),
and in all other images only when they are not matched by any predicted cell instance (\emph{FN}).
}
\label{fig:resultsA}
\vspace{-6pt}
\end{figure}
}}

\newcommand{\figResultsB}{{
\begin{figure}[t]
  \centering
  {\includegraphics[width=1\textwidth,clip,trim=10mm 8mm 2mm 2mm]{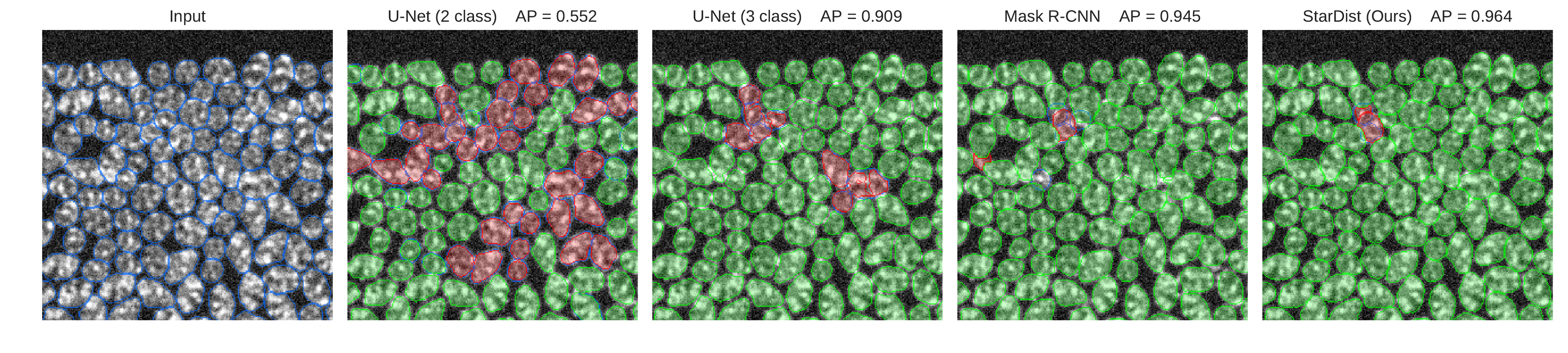}}%
\vspace{-8pt}
\caption{Segmentation result ($\tau=0.5$) for \Dtragen image. See \cref{fig:resultsA} caption for legend.}
\label{fig:resultsB}
\vspace{-6pt}
\end{figure}
}}

\newcommand{\figResultsCa}{{
\begin{figure}[t]
  \centering
  {\includegraphics[width=1\textwidth,clip,trim=10mm 8mm 2mm 2mm]{{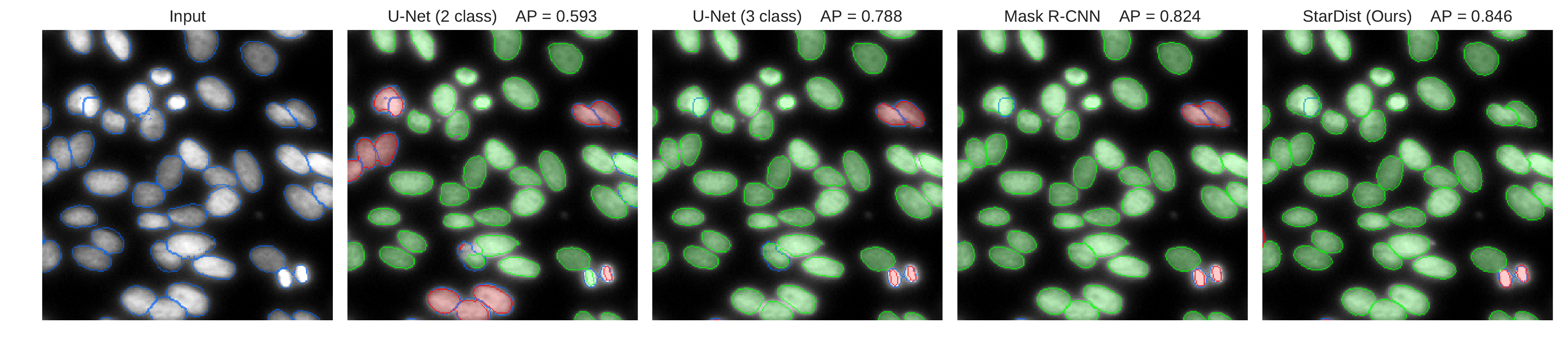}}}\\%
  {\includegraphics[width=1\textwidth,clip,trim=10mm 8mm 2mm 2mm]{{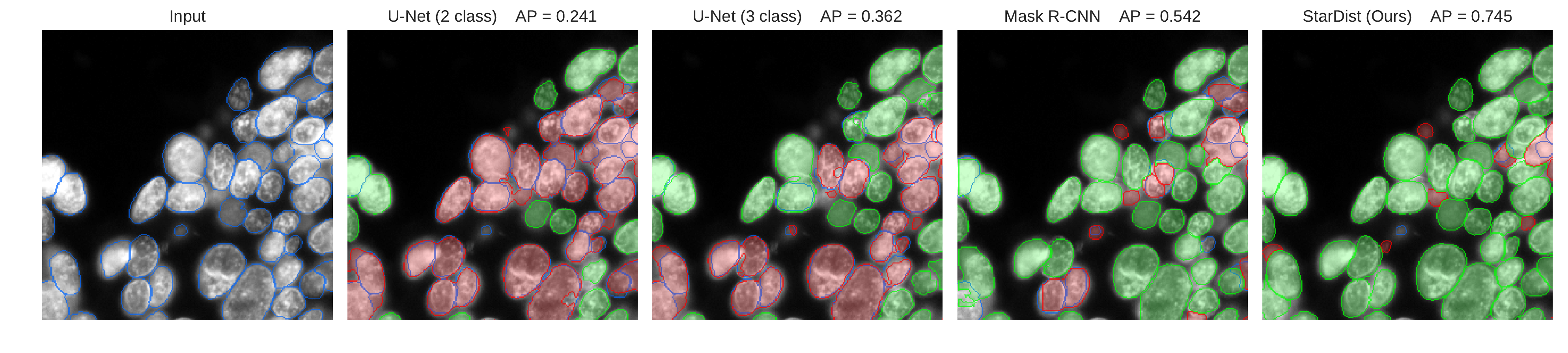}}}%
\vspace{-8pt}
\caption{Two segmentation results ($\tau=0.5$) for \Ddsb. See \cref{fig:resultsA} caption for legend.}
\label{fig:resultsCa}
\label{fig:resultsCb}
\vspace{-6pt}
\end{figure}
}}

\figOverview

\section{Introduction}
Many biological tasks rely on the accurate detection and segmentation of cells and nuclei from microscopy images~\cite{meijering2012cell}. Examples include high-content screens of variations in cell phenotypes \cite{boutros2015}, or the identification of developmental lineages of dividing cells \cite{amat2014,ulman2017}.
In many cases, the goal is to obtain an \emph{instance segmentation}, which is the assignment of a cell instance identity to every pixel of the image.
To that end, a prevalent \emph{bottom-up} approach is to first classify every pixel into semantic classes (such as \emph{cell} or \emph{background}) and then group pixels of the same class into individual instances. The first step is typically done with learned classifiers, such as random forests \cite{sommer2011ilastik} or neural networks \cite{ronneberger2015,chen2016dcan,pena2018}. Pixel grouping can for example be done by finding connected components \cite{chen2016dcan}. While this approach often gives good results, it is problematic for images of very crowded cell nuclei, since only a few mis-classified pixels can cause bordering but distinct cell instances to be fused \cite{caicedo2018,xie2018}.
An alternative \emph{top-down} approach is to first localize individual cell instances with a rough shape representation and then \emph{refine} the shape in an additional step. To that end, state-of-the-art object detection methods \cite{liu2016,redmon2016,ren2015} predominately predict axis-aligned bounding boxes, which can be refined to obtain an instance segmentation by classifying the pixels within each box (\eg, \emph{Mask R-CNN}~\cite{he2017}).
Most of these methods have in common that they avoid detecting the same object multiple times by performing a \emph{non-maximum suppression} (NMS) step where boxes with lower confidence are suppressed by boxes with higher confidence if they substantially overlap.
NMS can be problematic if the objects of interest are poorly represented by their axis-aligned bounding boxes, which can be the case for cell nuclei (\cref{fig:overview}a).
While this can be mitigated by using \emph{rotated} bounding boxes \cite{ma2018}, it is still necessary to refine the box shape to accurately describe objects such as cell nuclei.

To alleviate the aforementioned problems, we propose \stardist, a cell detection method that predicts a shape representation which is flexible enough such that -- without refinement -- the accuracy of the localization can compete with that of instance segmentation methods. To that end, we use \emph{star-convex polygons} that we find well-suited to approximate the typically roundish shapes of cell nuclei in microscopy images.
While Jetley et al. \cite{jetley2017} already investigated star-convex polygons for object detection in natural images, they found them to be inferior to more suitable shape representations for typical object classes in natural images, like people or bicycles.

In our experimental evaluation, we first show that methods based on axis-aligned bounding boxes (we choose Mask R-CNN as a popular example) cannot cope with certain shapes.
Secondly, we demonstrate that our method performs well on images with very crowded nuclei and does not suffer from merging bordering cell instances.
Finally, we show that our method exceeds the performance of strong competing methods on a challenging dataset of fluorescence microscopy images.
\stardist uses a light-weight neural network based on \emph{U-Net} \cite{ronneberger2015} and is easy to train and use, yet is competitive with state-of-art methods.
\section{Method}
\label{sec:method}

Our approach is similar to object detection methods \cite{redmon2016,liu2016,jetley2017} that directly predict shapes for each object of interest. %
Unlike most of them, we do not use axis-aligned bounding boxes as the shape representation (\cite{jetley2017,ma2018} being notable exceptions). Instead, our model predicts a \emph{star-convex polygon} for every pixel%
\footnote{Although we only consider the single object class \emph{cell nuclei} in our experiments,
note that we are not limited to that and thus use the generic term \emph{object} in the following.}.
Specifically, for each pixel with index $i,j$ we regress the distances $\{ r_{i,j}^k \}_{k=1}^n$ to the boundary of the object to which the pixel belongs, along a set of $n$ predefined radial directions with equidistant angles (\cref{fig:overview}b). %
Obviously, this is only well-defined for (non-background) pixels that are contained within an object.
Hence, our model also separately predicts for every pixel whether it is part of an object, so that we only consider polygon proposals from pixels with sufficiently high object probability $d_{i,j}$.
Given such polygon candidates with their associated object probabilities, we perform non-maximum suppression (NMS) to arrive at the final set of polygons, each representing an individual object instance.

\vspace{-6pt}
\paragraph{Object probabilities.}
While we could simply classify each pixel as either object or background based on binary masks, we instead define its object probability $d_{i,j}$ as the (normalized) Euclidean distance to the nearest background pixel (\cref{fig:overview}b). %
By doing this, NMS will favor polygons associated to pixels near the cell center (\cf\cref{fig:poly_example}b), which typically represent objects more accurately.

\vspace{-6pt}
\paragraph{Star-convex polygon distances.}
For every pixel belonging to an object, the Euclidean distances $r_{i,j}^k$ to the object boundary can be computed by simply following each radial direction $k$ until a pixel with a different object identity is encountered.
We use a simple GPU implementation that is fast enough that we can compute the required distances on demand during model training.

\subsection{Implementation}

Although our general approach is not tied to a particular regression or classification approach, we choose the popular U-Net \cite{ronneberger2015} network as the basis of our model.
After the final U-Net feature layer, we cautiously add an additional $3{\times}3$ convolutional layer with $128$ channels (and \emph{relu} activations) to avoid that the subsequent two output layers have to ``fight over features''.
Specifically, we use a single-channel convolutional layer with \emph{sigmoid} activation for the object probability output.
The polygon distance output layer has as many channels as there are radial directions $n$ and does not use an additional activation function.

\vspace{-6pt}
\paragraph{Training.}
We minimize a standard \emph{binary cross-entropy} loss for the predicted object probabilities.
For the polygon distances, we use a \emph{mean absolute error} loss weighted by the ground truth object probabilities, \ie the pixel-wise errors are multiplied by the object probabilities before averaging.
Consequently, background pixels will not contribute to the loss, since their object probability is zero.
Furthermore, predictions for pixels closer to the center of each object are weighted more, which is appropriate since these will be favored during non-maximum suppression.
The code is publicly available\footnote{\url{https://github.com/mpicbg-csbd/stardist}}.

\vspace{-6pt}
\paragraph{Non-maximum suppression.}

We perform common, greedy non-maximum suppression (NMS, \cf\cite{ren2015,liu2016,redmon2016}) to only retain those polygons in a certain region with the highest object probabilities.
We only consider polygons associated with pixels above an object probability threshold as candidates, and compute their intersections with a standard polygon clipping method.

\section{Experiments}

\figResultsA

\subsection{Datasets}
We use three datasets that pose different challenges for cell detection:

\vspace{-6pt}
\paragraph{Dataset \Dtoy:} Synthetically created images that contain pairs of touching half-ellipses with blur and background noise (\cf~\cref{fig:resultsA}).
Each pair is oriented in such a way that the overlap of both enclosing bounding boxes is either very small (along an axis-aligned direction) or very large (when the ellipses touch at an oblique angle).
This dataset contains $1000$ images of size $256\times 256$ with associated ground truth labels. We specifically created this dataset to highlight the limitations of methods that predict axis-aligned bounding boxes.

\vspace{-6pt}
\paragraph{Dataset \Dtragen:} Synthetically generated images of an evolving cell population from \cite{ulman2015} (\cf~\cref{fig:resultsB}). The generative model includes cell divisions, shape deformations, camera noise and microscope blur and is able to simulate realistic images of extremely crowded cell configurations. This dataset contains $200$ images of size $792\times 792$ along with their ground truth labels.

\vspace{-6pt}
\paragraph{Dataset \Ddsb:} Manually annotated real microscopy images of cell nuclei from the 2018 Data Science Bowl\footnote{\url{https://www.kaggle.com/c/data-science-bowl-2018}}. From the original dataset ($670$ images from diverse modalities) we selected a subset of fluorescence microscopy images and removed images with labeling errors, yielding a total of $497$ images (\cf~\cref{fig:resultsCa}).

\vspace{3pt}
For each dataset, we use $90\%$ of the images for training and $10\%$ for testing.
We train all methods (\cref{sub:compared_methods}) with the same random crops of size $256\times 256$ from the training images (augmented via axis-aligned rotations and flips).

\subsection{Evaluation Metric}

We adopt a typical metric for object detection: A detected object $\mathit{I}_{\text{pred}}$ is considered a match (\emph{true positive} $\mathit{TP}_\tau$) if a ground truth object $\mathit{I}_{\text{gt}}$ exists whose \emph{intersection over union} $\mathit{IoU} = \frac{\mathit{I}_{\text{pred}} \cap \mathit{I}_{\text{gt}}}{\mathit{I}_{\text{pred}} \cup \mathit{I}_{\text{gt}}}$ is greater than a given threshold $\tau \in [0,1]$. Unmatched predicted objects are counted as \emph{false positives} ($\mathit{FP}_\tau$), unmatched ground truth objects as \emph{false negatives} ($\mathit{FN}_\tau$).
We use the \emph{average precision} $\mathit{AP}_\tau = \frac{\mathit{TP}_\tau}{\mathit{TP}_\tau+\mathit{FN}_\tau+\mathit{FP}_\tau}$ evaluated across all images as the final score.

\figResultsB

\subsection{Compared methods}
\label{sub:compared_methods}

\paragraph{U-Net (2 class):}
We use the popular U-Net architecture~\cite{ronneberger2015} as a baseline to predict $2$ output classes (cell, background). We use $3$ down/up-sampling blocks, each consisting of $2$ convolutional layers with $32\cdot 2^k (k = 0,1,2)$ filters of size $3\times 3$ (approx. $1.4$ million parameters in total). We apply a threshold $\sigma$ on the cell probability map and retain the connected components as final result ($\sigma$ is optimized on the validation set for every dataset).

\vspace{-6pt}
\paragraph{U-Net (3 class):} Like U-Net (2 class), but we additionally predict the \emph{boundary pixels} of cells as an extra class. The purpose of this is to differentiate crowded cells with touching borders (similar to \cite{chen2016dcan,pena2018}). We again use the connected components of the thresholded cell class as final result.

\vspace{-6pt}
\paragraph{Mask R-CNN:} A state-of-the-art instance segmentation method combining a bounding-box based region proposal network, non-maximum-suppression (NMS), and a final mask segmentation (approx. 45 million parameters in total). We use a popular open-source implementation\footnote{\url{https://github.com/matterport/Mask_RCNN}}. For each dataset, we perform a grid-search over common hyper-parameters, such as detection NMS threshold, region proposal NMS threshold, and number of anchors.

\vspace{-6pt}
\paragraph{\stardist:} Our proposed method as described in \cref{sec:method}. We always use $n=32$ radial directions (\cf\cref{fig:overview}b) and employ the same U-Net backbone as for the first two baselines described above.

\figResultsCa

\subsection{Results}

We first test our approach on dataset \Dtoy, which was intentionally designed to contain objects with many overlapping bounding boxes.
The results in \cref{tab:results} and \cref{fig:resultsA} show that
for moderate IoU thresholds ($\tau < 0.7$), \stardist and both U-Net baselines yield essentially perfect results. %
Mask R-CNN performs substantially worse due to the presence of many slanted and touching pairs of objects (which have almost identical bounding boxes, hence one is suppressed).
This experiment highlights a fundamental limitation of object detection methods that predict axis-aligned bounding boxes.

On dataset \Dtragen, U-Net (2 class) shows the lowest accuracy mainly due to the abundance of touching cells which are erroneously fused. \cref{tab:results} shows that all other methods attain almost perfect accuracy for many IoU thresholds even on very crowded images, which might be due to the stereotypical size and texture of the simulated cells.
We show the most difficult test image in \cref{fig:resultsB}. %

\figPolyExample

Finally, we turn to the real dataset \Ddsb where we find \stardist to outperform all other methods for IoU thresholds $\tau < 0.75$, followed by the next best method Mask R-CNN (\cf~\cref{tab:results} and \cref{fig:poly_example}a).
\cref{fig:resultsCa} shows the results and errors for two different types of cells. Common segmentation errors include merged cells (mostly for the 2 class U-Net), bounding box artifacts (Mask R-CNN) and missing cells (all methods).
The bottom example of \cref{fig:resultsCb} is particularly challenging, where out-of-focus signal results in densely packed and partially overlapping cell shapes. Here, merging mistakes are pronounced for both U-Net baselines.
All false positives predicted by \stardist retain a reasonable shape, whereas those predicted by Mask R-CNN sometimes exhibit obvious artifacts.

We observe that \stardist yields inferior results for the largest IoU thresholds $\tau$ for our synthetic datasets.
This is not surprising, since we predict a parametric shape model based on only $32$ radial directions,
instead of a per-pixel segmentation as all other methods.
However, an advantage of a parametric shape model is that it can be used to predict reasonable complete shape hypotheses from nuclei that are only partially visible at the image boundary (\cf~\cref{fig:poly_example}b, also see \cite{yurchenko2017}).

\begin{table}[t]
  \centering
  \tabcolsep=3.45pt
  \scriptsize
  \begin{tabular}{@{} l S[table-format=1.4]S[table-format=1.4]S[table-format=1.4]S[table-format=1.4]S[table-format=1.4]S[table-format=1.4]S[table-format=1.4]S[table-format=1.4]S[table-format=1.4] @{}}
    \toprule
    Threshold $\tau$ & {0.50} & {0.55} & {0.60} & {0.65} & {0.70} & {0.75} & {0.80} & {0.85} & {0.90} \\
    \midrule
    \multicolumn{10}{c}{{\Dtoy}}\\ %
    U-Net (2 class) & 0.99937 & 0.99896 & 0.99771 & 0.99314 & 0.96414 & 0.86588 & 0.62293 & 0.29387 & 0.06666 \\ %
    U-Net (3 class) & \bfseries 0.99979 & \bfseries 0.99979 & \bfseries 0.99979 & \bfseries 0.99979 & \bfseries 0.99979 & \bfseries 0.99979 & \bfseries 0.99896 & \bfseries 0.98736 & \bfseries 0.92434 \\ %
    Mask R-CNN      & 0.91042 & 0.90610 & 0.90140 & 0.89439 & 0.87293 & 0.84714 & 0.77284 & 0.60750 & 0.37167 \\ %
    StarDist (Ours) & \bfseries 0.99979 & \bfseries 0.99979 & \bfseries 0.99979 & \bfseries 0.99979 & 0.99937 & 0.98900 & 0.86954 & 0.46299 & 0.07478 \\ %
    \midrule
    \multicolumn{10}{c}{{\Dtragen}}\\
    U-Net (2 class) & 0.90297 & 0.89077 & 0.88516 & 0.88151 & 0.88108 & 0.87830 & 0.85655 & 0.69374 & 0.40557 \\ %
    U-Net (3 class) & 0.99177 & 0.99036 & 0.98990 & 0.98966 & 0.98896 & 0.98826 & \bfseries 0.98477 & \bfseries 0.96793 & \bfseries 0.89951 \\ %
    Mask R-CNN      & 0.99236 & 0.99189 & 0.99118 & 0.98978 & 0.98628 & 0.97770 & 0.95940 & 0.89475 & 0.52801 \\ %
    StarDist (Ours) & \bfseries 0.99835 & \bfseries 0.99811 & \bfseries 0.99764 & \bfseries 0.99670 & \bfseries 0.99529 & \bfseries 0.99342 & 0.98409 & 0.94652 & 0.42593 \\ %
    \midrule
    \multicolumn{10}{c}{{\Ddsb}}\\
    U-Net (2 class) & 0.67391 & 0.62952 & 0.59747 & 0.56504 & 0.53390 & 0.48192 & 0.41506 & 0.32479 & 0.20317 \\ %
    U-Net (3 class) & 0.80602 & 0.77531 & 0.74311 & 0.70113 & 0.65427 & 0.57767 & \bfseries 0.49100 & \bfseries 0.37375 & \bfseries 0.22582 \\ %
    Mask R-CNN      & 0.83227 & 0.80505 & 0.77279 & 0.72989 & 0.68375 & \bfseries 0.59744 & 0.48928 & 0.35253 & 0.18909 \\ %
    StarDist (Ours) & \bfseries 0.86405 & \bfseries 0.83608 & \bfseries 0.80428 & \bfseries 0.75448 & \bfseries 0.68503 & 0.58620 & 0.44951 & 0.28647 & 0.11911 \\ %
    \bottomrule
  \end{tabular}
  \vspace{2pt}
  \caption{Cell detection results for three datasets and four methods, showing \emph{average precision} (AP) for several \emph{intersection over union} (IoU) thresholds $\tau$.}
  \label{tab:results}
  \vspace{-6pt}
\end{table}

\section{Discussion}

We demonstrated that star-convex polygons are a good shape representation to accurately localize cell nuclei even under challenging conditions. Our approach is especially appealing for images of very crowded cells.
When our \stardist model makes a mistake, it does so gracefully by either simply omitting a cell or by predicting at least a plausible cell shape. The same cannot by said for the methods that we compared to, whose predicted shapes are sometimes obviously implausible (\eg, containing holes or ridges).
While \stardist is competitive to the state-of-the-art Mask R-CNN method, a key advantage is that it has an order of magnitude fewer parameters and is much simpler to train and use.
In contrast to Mask R-CNN, \stardist has only few hyper-parameters that do not need careful tuning to achieve good results.

Our approach could be particularly beneficial in the context of cell tracking. There, it is often desirable to have multiple diverse segmentation hypotheses~\cite{rempfler2017,jug2016}, which could be achieved by suppressing fewer candidate polygons. Furthermore, \stardist can plausibly complete shapes for partially visible cells at the image boundary, which could make it easier to track cells that enter and leave the field of view over time.

\bibliographystyle{splncs04}
% \bibliography{literature}

\end{document}